%% file: main.tex
\documentclass[sigconf]{acmart}

\usepackage{algorithm}
\usepackage{algorithmic}
\usepackage{xspace}
\usepackage{multirow}
\usepackage{multicol}
\usepackage{booktabs}
\usepackage{graphicx}  
\usepackage{float}  
\usepackage{subfigure}  
\usepackage{pifont}
\usepackage{makecell}
\usepackage{tabularx}

\newtheorem{myDef}{Definition}
\newtheorem{prob}{Problem}

\newcommand{\etal}{\emph{et~al.}}
\newcommand{\eg}{\emph{e.g.},\/~}

\newcommand{\ie}{\emph{i.e.},\/~}
\newcommand{\wrt}{\emph{w.r.t.}\/~}

\newcommand\figref[1]{Figure~\ref{#1}}

\newcommand\tabref[1]{Table~\ref{#1}}
\newcommand\secref[1]{Section~\ref{#1}}
\newcommand\appref[1]{Appendix~\ref{#1}}
\newcommand{\model}{{\sc MixTTE}\xspace}

\newcommand{\eat}[1]{}

\newcommand{\TODO}[1]{{\color{red}TODO: {#1}}}
\newcommand\beftext[1]{{\color[rgb]{0.5,0.5,0.5}{BEFORE:#1}}}
\newcommand{\rev}[1]{{\color{purple}#1}} 
\newcommand{\wz}[1]{{\color[rgb]{0.5,0.7,0.4}{#1}}}

\AtBeginDocument{%
  }

\setcopyright{acmlicensed}
\copyrightyear{2026}
\acmYear{2026}
\setcopyright{cc}
\setcctype{by}
\acmConference[KDD '26]{Proceedings of the 32nd ACM SIGKDD Conference on Knowledge Discovery and Data Mining V.1}{August 09--13, 2026}{Jeju Island, Republic of Korea}
\acmBooktitle{Proceedings of the 32nd ACM SIGKDD Conference on Knowledge Discovery and Data Mining V.1 (KDD '26), August 09--13, 2026, Jeju Island, Republic of Korea}
\acmPrice{}
\acmDOI{10.1145/3770854.3783940}
\acmISBN{979-8-4007-2258-5/2026/08}

\begin{document}



\title{A Scalable and Adaptive Multi-level Route-link Mixture Framework for Travel Time Estimation on Million-Scale Road Networks}
\title{RouLiTTE: A Scalable and Adaptive Multi-level Route-link Mixture Framework for Large-scale Travel Time Estimation}
\title{MixTTE: Multi-Level Mixture-of-Experts for Scalable and Adaptive Travel Time Estimation}


\author{Wenzhao Jiang}
\orcid{0009-0006-1081-8684}
\affiliation{
\institution{The Hong Kong University of Science and Technology (Guangzhou)}
\city{Guangzhou}
\state{Guangdong}
\country{China}
}
\email{wjiang431@connect.hkust-gz.edu.cn}

\author{Jindong Han}
\orcid{0000-0002-1542-6149}
\affiliation{
\institution{The Hong Kong University of Science and Technology}
\city{Hong Kong}
\country{China}
}
\email{jhanao@connect.ust.hk}

\author{Ruiqian Han}
\orcid{0009-0007-0440-735X}
\affiliation{
\institution{The Hong Kong University of Science and Technology (Guangzhou)}
\city{Guangzhou}
\state{Guangdong}
\country{China}
}
\email{rhan464@connect.hkust-gz.edu.cn}

\author{Hao Liu}
\authornote{Corresponding author.}
\orcid{0000-0003-4271-1567}
\affiliation{
\institution{The Hong Kong University of Science and Technology (Guangzhou)}
\city{Guangzhou}
\state{Guangdong}
\country{China}
}
\affiliation{
\institution{The Hong Kong University of Science and Technology}
\city{Hong Kong}
\country{China}
}
\email{liuh@ust.hk}







\renewcommand{\shortauthors}{Wenzhao Jiang, Jindong Han, Ruiqian Han, and Hao Liu}

\begin{abstract}

Accurate Travel Time Estimation (TTE) is critical for ride-hailing platforms, where errors directly impact user experience and operational efficiency. 
While existing production systems excel at holistic route-level dependency modeling, they struggle to capture city-scale traffic dynamics and long-tail scenarios, leading to unreliable predictions in large urban networks.
In this paper, we propose \model, a scalable and adaptive framework that synergistically integrates link-level modeling with industrial route-level TTE systems. 
Specifically, we propose a spatio-temporal external attention module to capture global traffic dynamic dependencies across million-scale road networks efficiently.
Moreover, we construct a stabilized graph mixture-of-experts network to handle heterogeneous traffic patterns while maintaining inference efficiency.
Furthermore, an asynchronous incremental learning strategy is tailored to enable real-time and stable adaptation to dynamic traffic distribution shifts.
Experiments on real-world datasets validate \model significantly reduces prediction errors compared to seven baselines.
\model has been deployed in DiDi, substantially improving the accuracy and stability of the TTE service.
\end{abstract}

\ccsdesc[500]{Computing methodologies~Machine learning}
\ccsdesc[500]{Applied computing~Transportation}
\ccsdesc[500]{Information systems~Spatial-temporal systems}

\keywords{travel time estimation, external attention, mixture-of-experts, asynchronous incremental learning} 



\maketitle

\input{introduction.tex}
\input{preliminary.tex}

\input{method.tex}
\input{deployment.tex}

\input{experiments.tex}

\input{relatedwork.tex}
\input{conclusion.tex}


\bibliographystyle{ACM-Reference-Format}
\balance
\bibliography{sample}

\appendix

\section{Methodology Details}
\subsection{Intra-time Spatial Hierarchical Modeling within STEA}
\label{app:intra_time}
Despite the capacity of external memory units in bridging the interactions among traffic patterns across both space and time, it is still worthwhile to reinforce the intra-time step global spatial correlation modeling for practical appliacations.
As briefed in~\secref{subsec:stea} and \figref{fig:overall_framework}, we adopt a spatial hierarchical modeling method~\cite{sshgnn2022tkde,transolver2024icml} to achieve this goal. By projecting fine-grained link-level traffic contexts into semantic tokens, we can more efficiently model the global spatial correlation while maintaining the seamless integration with the EKR step.
Here we provide more details on the soft clustering and broadcasting steps that are essential to enable the hierarchical modeling.
Given the link representations $\mathbf{H} \in \mathbb{R}^{N \times d}$, the soft clustering step leverage a learnable weight matrix $\mathbf{W}_{sc} \in \mathbb{R}^{d \times U_{in}}$ followed by a softmax activation to produce the soft clustering weights for each link, \ie 
\begin{align}
  \alpha_i = \operatorname{Softmax}\left(\mathbf{h}_i \cdot \mathbf{W}_{sc}\right) \in \mathbb{R}^{U_{in}}, \quad \forall i = 1, \ldots, N.
\end{align}
Then, we obtain the traffic semantic tokens by aggregating the link representations according to the soft clustering weights, \ie
\begin{align}
  \mathbf{h}^{se}_i = \sum_{j=1}^{N} \alpha_{ij} \cdot \mathbf{h}_j \mathbf{W}_{proj} \in \mathbb{R}^{d}, \quad \forall i = 1, \ldots, U_{in},
\end{align}
where $\mathbf{W}_{proj} \in \mathbb{R}^{d \times d}$ is a learnable weight matrix to project the traffic slice encodings into a unified semantic space.
After performing self-attention and EKR steps on these semantic tokens, we will obtain the externally augmented semantic tokens $\mathbf{H}^{se,ex} \in \mathbb{R}^{U_{in} \times d}$ that absorb both the intra-time and inter-time global traffic correlations.
Finally, we BroadCast~(BC) back these semantic tokens to produce the externally augmented link representations according to the soft clustering weights, \ie
\begin{align}
  \mathbf{h}^{ex}_i = \sum_{j=1}^{U_{in}} \alpha_{ij} \cdot \mathbf{h}^{se,ex}_j \in \mathbb{R}^{d}, \quad \forall i = 1, \ldots, N.
\end{align}


\section{Experimental Details} \label{app:experiment}

\subsection{Hot Link Selection} \label{app:hotlink}
On a daily basis, we generate the hot degree feature of each link by jointly considering the link's congestion levels and the number of trips passing through it. The links that surpass a predefined hot degree are considered 'hot links'.
Then, on a weekly basis, we union the hot links over the past month to form the target hot link set for the next week's link-level model training.
In practice, we also include the neighboring links of the selected hot links to prevent them from being isolated in the traffic network.
When dealing with million-scale urban road networks, this process is not only efficient but also useful for identifying the links that exhibit prominent traffic patterns, which are supposed to be the key focus of link-level modeling to provide tangible traffic contextualization for the route-centric TTE system.

\subsection{Implementation Details}
\label{app:imp_detail}
We provide the implementation details of \model in this section.
The hidden dimension of \model set to 64 for the link-level model and 256 for the route-centric model.
In the STEA module, the number of external memory units is 128 with each memory unit having a dimension of 64. We also leverage multi-head attention with 8 heads in both EKR and self-attention steps.
In the ESGMoE module, the total layers of ESGMoE is 2, with each layer consisting of 16 experts, including 12 graph experts, 1 zero experts, 1 identity expert and 2 constant expert. Each graph expert has an 8-hop receptive field. The top-2 experts will be activated at a time. 
The coefficient of the expert load balancing loss is set to 1000 for Beijing, and 100 for Nanjing and Suzhou. The weight of zero-computation experts is set to 0.5 for Beijing, and 1.5 for Nanjing and Suzhou.
In the ASIL module, the percentile for detecting an anomalous link is set to 0.75. The percentile for activating the link-level model update and both the route-level and link-level model update is set to 0.75 and 0.9, respectively.  
We use Adam Optimization with learning rate $10^{-3}$, weight decay rate $5\times10^{-7}$, dropout rate 0.3 and 2 epochs for training.






\subsection{Hyperparameter Sensitivity Analysis}
\begin{figure}[t] 
\begin{minipage}{1.0\linewidth}
\centering  
\subfigure[\small Effect of $(\delta_d, \delta_l, \delta_r)$.]{
    \label{fig:asil}
    \includegraphics[width=0.51\linewidth]{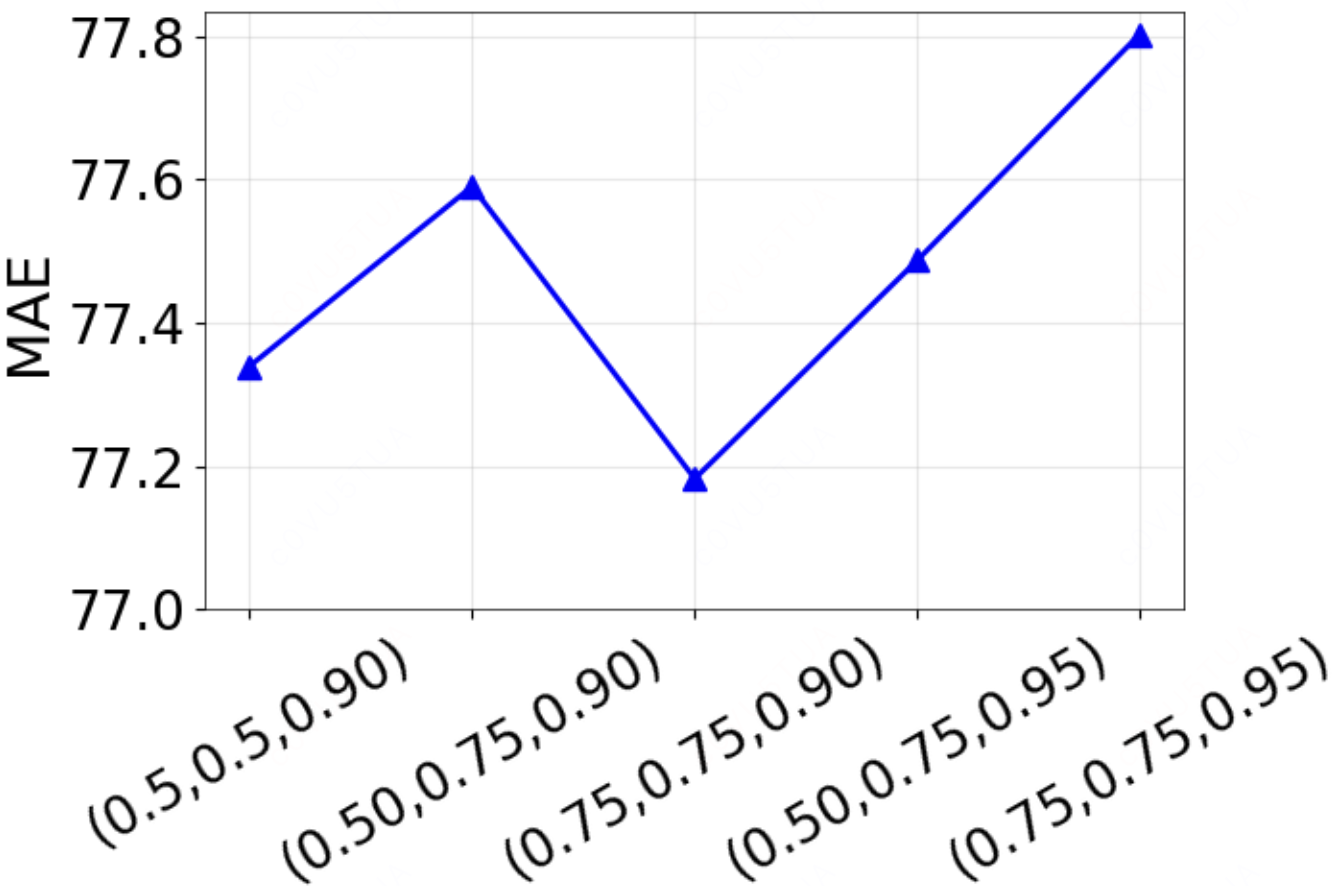}}
\subfigure[\small Effect of $\delta$.]{
    \label{fig:delta}
    \includegraphics[width=0.45\linewidth]{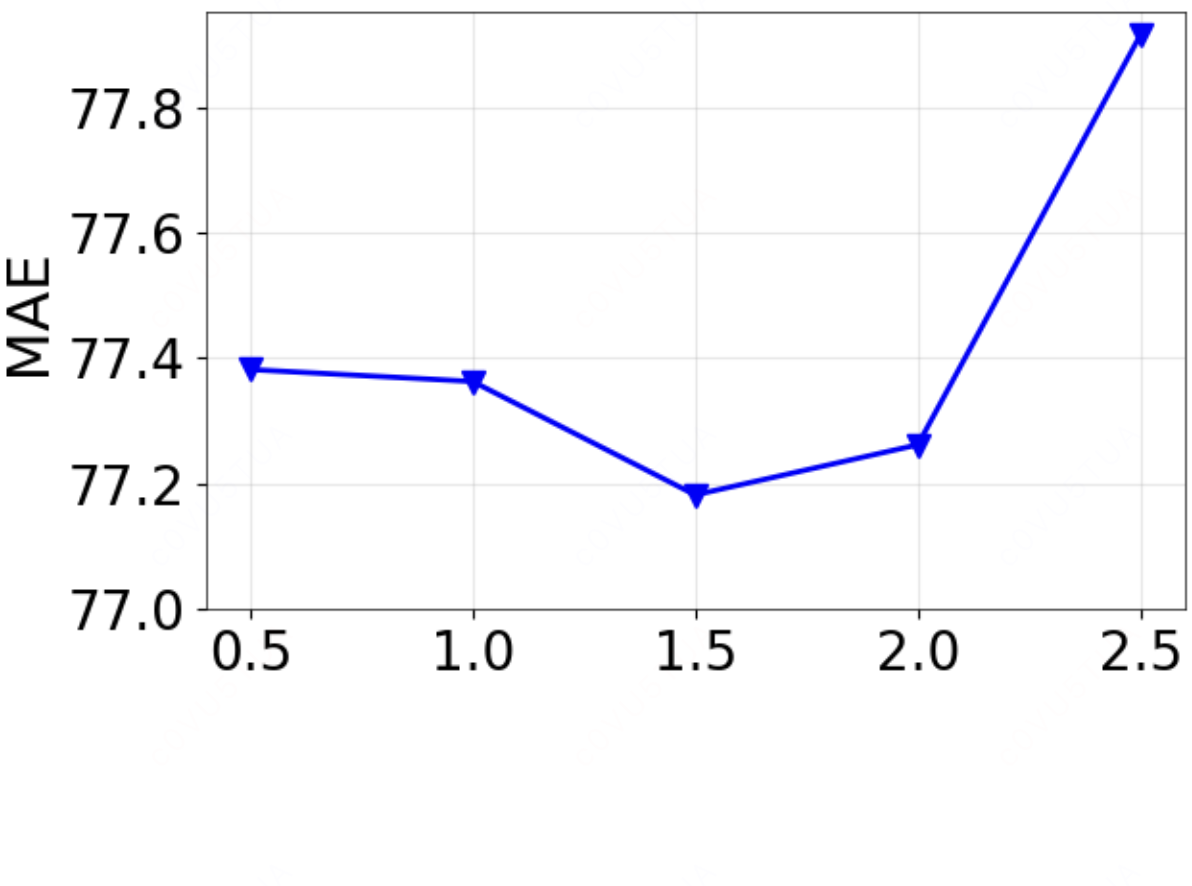}}
\\
\subfigure[\small Effect of $N_{graph}$.]{
    \label{fig:N_graph}
    \includegraphics[width=0.45\linewidth]{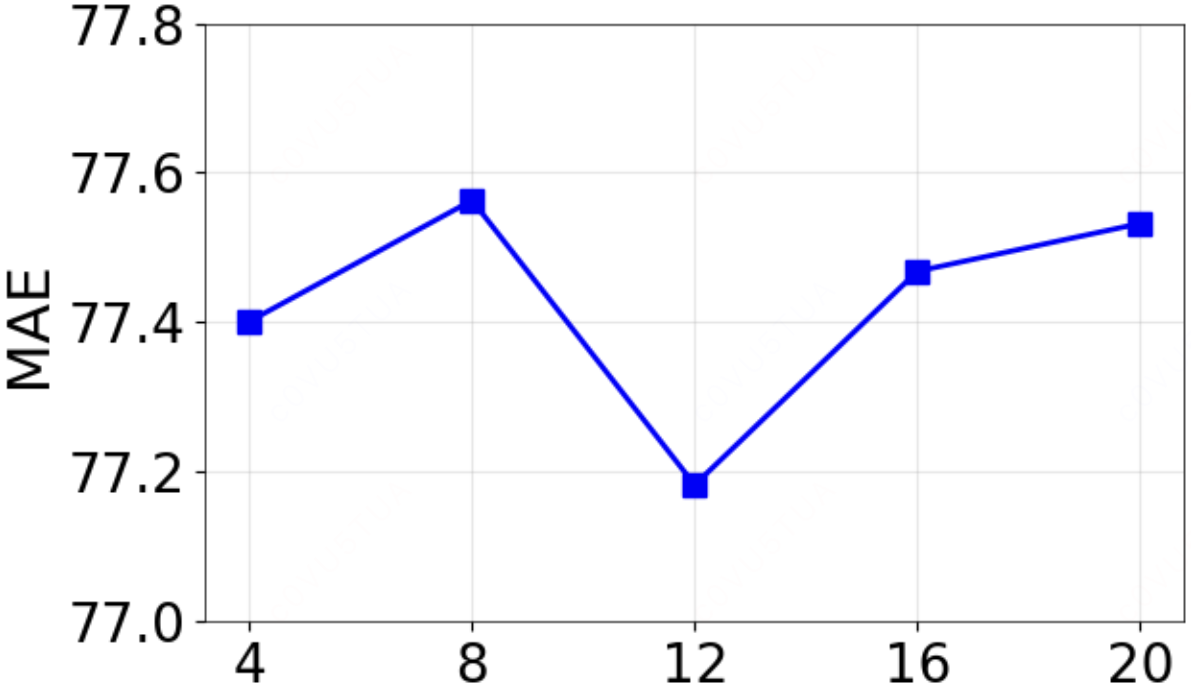}}
\quad
\subfigure[\small Effect of $U_{ex}$.]{
    \label{fig:U_ex}
    \includegraphics[width=0.45\linewidth]{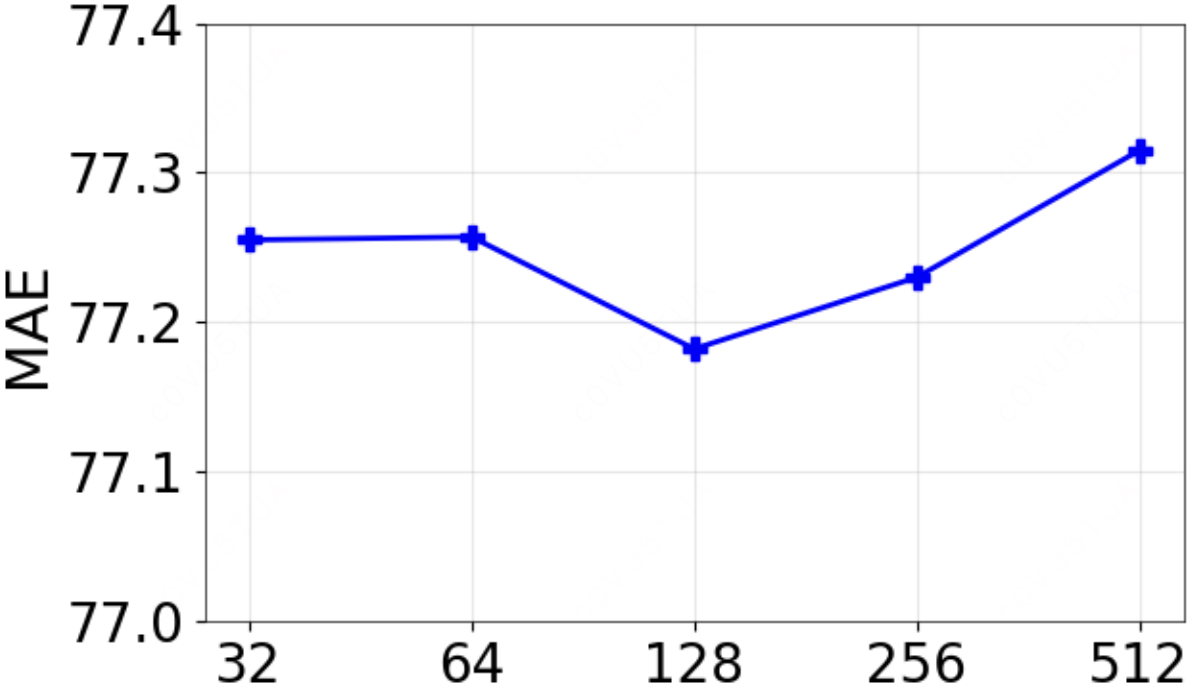}} 
\end{minipage}
\caption{Parameter sensitivity analysis in Suzhou dataset.}
\label{fig:sens}
\end{figure}

We test the sensitivity of \model on Suzhou dataset \wrt four hyperparameters: (1) thresholds in ASIL $(\delta_d, \delta_l, \delta_r)$, (2) load balancing weight for zero-computation experts ($\delta$), (3) number of graph experts ($N_{graph}$) and (4) number of external units ($U_{ex}$). The performance variations \wrt the MAE metric are shown in \figref{fig:sens}.
Overall, MixTTE's performances vary within an acceptable range, demonstrating its robustness against these hyperparameters. Specifically, the performance experiences an up and down \wrt the change of each hyperparameter. For $(\delta_d, \delta_l, \delta_r), $ low thresholds lead to instability from noise, while high thresholds delay adaptation to real traffic changes. For $\delta$, a small value may misroute complex patterns to zero-computation experts, whereas a large value may waste graph experts on frequent patterns. For $N_{graph}$, too few results in a model with insufficient capacity, while too many leads to undertraining and poor specialization. For $U_{ex}$, an insufficient number limits the model's global context awareness, while an excessive number increases the risk of overfitting.

\section{Future Work}
Looking ahead, we aim to evolve MixTTE into a more comprehensive foundational framework by expanding its capabilities in three key dimensions. 
First, we plan to incorporate multi-modal data, such as weather conditions and social events, to further mitigate the partial observability issue in urban scenarios and enhance the model's robustness and generalizability.
The STEA module's modality-agnostic nature can naturally support integrating these multi-modal factors into a unified embedding space, forming more comprehensive memory banks. 
Second, we will investigate cross-city transfer learning techniques to generalize traffic patterns across different urban networks, significantly reducing the calibration costs required for deploying the system in new cities. The external memory and the heterogeneous MoE orchestration have laid the architecture foundation for abstracting more cross-city knowledge.
Finally, we intend to explore lightweight model distillation techniques to facilitate the deployment of large-scale cross-city, cross-modal foundation model we plan to construct.








\end{document}

%% file: introduction.tex
\section{Introduction}
Travel Time Estimation (TTE) aims to predict the duration of a given travel route that consists of an ordered sequence of road links connecting an origin to a destination.
Modern ride-hailing platforms like DiDi process over billions of travel time queries daily to power essential services, including route planning, order dispatching, and dynamic pricing. 
With massive active users across numerous cities, the operational efficiency of the platform hinges on the accuracy and reliability of its Travel Time Estimation~(TTE) system, where even tiny improvements translate to millions in annual savings and significantly enhanced user satisfaction.

The current production system at DiDi employs a \textbf{route-centric} Wide-Deep-Recurrent (WDR) architecture~\cite{wdr2018kdd} to capture intricate route feature interactions and sequential dependencies among successive in-route links.
Over the past years, DiDi continuously advances the feature engineering, model architecture and training strategy~\cite{ieta2023kdd,probtte2023kdd} to enhance the accuracy, efficiency and robustness of the TTE system.
While effective for common scenarios, this approach suffers from two critical limitations:
\textit{(1)~Limited reception field}. By treating routes as isolated sequences, the system fails to account for broader traffic dynamics that propagate into routes (\eg congestion spreading from nearby highways).
\textit{(2)~Long-tail underperformance}. The monolithic architecture struggles with rare but critical patterns (\eg event traffic, construction zones), which induce higher error rates on tail scenarios.
In this work, we aim to improve the current TTE system by integrating \textbf{link-level} capability that explicitly captures complex spatio-temporal dependencies across both in-route and off-route links for stronger awareness of contextual traffic conditions.

\begin{figure}[t] 
\centering  
\subfigure[Trip duration.]{ 
    \includegraphics[width=0.31\linewidth]{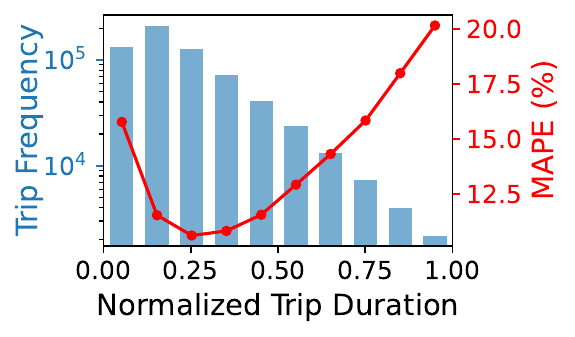}}
\subfigure[En-route traffic condition deviation degree.]{
    \includegraphics[width=0.31\linewidth]{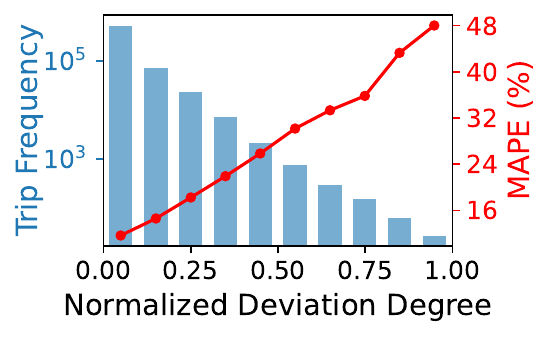}}
\subfigure[En-route traffic condition non-recurrence degree.]{
    \includegraphics[width=0.31\linewidth]{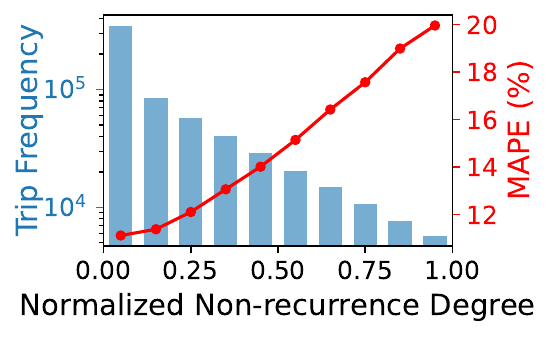}}
\caption{ 
The long-tail distributions of ride-hailing trip data \wrt three route-centric measurements. 
A state-of-the-art route-centric TTE method~\cite{ieta2023kdd} still performs poorly on a non-negligible fraction of tail routes.
}
\label{fig:longtail}
\end{figure}

Despite considerable research on link-level TTE algorithms~\cite{compacteta2020kdd,constgat2020kdd,googleeta2021cikm,sthr2022vldb,dueta2022cikm}, integrating them with industrial route-centric systems poses three primary challenges rooted in scalability, heterogeneity, and dynamic adaptation.
First, it remains inefficient for \textit{modeling global spatio-temporal dependencies} on large-scale road networks. Urban traffic shows long-range correlations (\eg residential-area morning congestion propagating to business districts hours later) and semantically similar but temporally distant patterns (\eg morning and evening rush hours). 
While recent work~\cite{dueta2022cikm,sthr2022vldb} improves local context modeling, efficient and effective global route context awareness remains under-explored in large-scale TTE. 
Existing methods~\cite{agcrn2020nips,staeformer2023cikm} that attempt global modeling typically rely on pairwise link correlations with quadratic complexity, rendering them impractical for million-scale road networks.
Second, \textit{traffic pattern heterogeneity} introduces long-tail TTE accuracy bottlenecks. As depicted in~\figref{fig:longtail}, urban road networks exhibit diverse dynamics driven by zoning, time, and exogenous factors~(\eg weather, events), creating skewed distributions where rare but critical scenarios (\eg stadium events) are poorly handled. Simply scaling model capacity risks overfitting frequent patterns while neglecting tail generalization, as monolithic architectures conflate unrelated patterns during joint optimization. 
Recently, Mixture-of-Experts~(MoE) architectures~\cite{testam2024iclr,cpmoe2024kdd} show promise in handling divergent patterns via conditional computation and sparse expert activation. However, vanilla MoE solutions~\cite{sparsemoe2017iclr,switch2022jmlr} lack stabilization mechanisms for noisy, graph-structured traffic data, leading to expert collapse or interference, undermining their effectiveness for TTE.
Third, \textit{continuous adaptation to traffic shifts} becomes computationally intractable with naive integration. While incremental learning~(IL) has proven cost-effective in route-centric systems~\cite{ieta2023kdd}, adding link-level modeling exacerbates two bottlenecks: (1) Parameter explosion from fine-grained link representations makes frequent full-model updates infeasible under latency budgets; (2) Increased sensitivity to transient link-level shifts risks overfitting noise or forgetting stable route-level patterns. 
Current IL strategies~\cite{ieta2023kdd,ctpugr2023icdm} lack modularity to decouple update frequencies for multi-level trip components~(\ie high-frequency link updates v.s. low-frequency route updates), hindering cost-effective adaptation.

In this work, we propose \model, a scalable and adaptive multi-level TTE framework that modularly integrates link-level modeling into DiDi's current route-centric system.
First, we propose a spatio-temporal external attention module to efficiently capture global dependencies across million-scale road networks, thereby overcoming the scalability bottleneck of link representation and enriching its traffic context awareness.
Second, we construct an externally stabilized graph Mixture-of-Experts~(MoE) module to facilitate the model to handle long-tail scenarios by mitigating interference among diverse heterogeneous traffic patterns.
Finally, we develop an asynchronous incremental learning strategy that selectively updates route- and link-level model parameters in response to detected distribution shifts, which allows for real-time adaptation with sustainable computational costs.
Notably, \model requires no refactoring of the existing data pipeline or route-centric model, allowing seamless plug-in integration with DiDi's existing production system.
Since April 2025, \model has been deployed on DiDi's ride-hailing platform, significantly improving the service's effectiveness and user experience.


Our major contributions are summarized below:
(1) We propose \model, a scalable and adaptive framework that synergistically integrates link-level and route-level model advances into DiDi's TTE system.
(2) We introduce a spatio-temporal external attention module and an externally stabilized graph MoE to efficiently capture global spatio-temporal dependencies and accommodate fine-grained heterogeneous patterns, respectively.
(3) We tailor an asynchronous incremental learning strategy for the route-link mixture model update, enabling efficient and stable real-time adaptation to traffic distribution shifts.
(4) Extensive offline and online experiments demonstrate superior accuracy, scalability, and adaptability over state-of-the-art approaches.
We also share insights from real-world deployments to support future industrial adoption.

%% file: preliminary.tex
\section{Preliminary}

\subsection{Definitions and Problem Statement}
This paper focuses on estimating ride-hailing travel times within the road network. We first present key definitions as follows.

\begin{myDef}[Traffic Network]
    A traffic network is modeled as a directed weighted graph $\mathcal{G}=(\mathcal{V},\mathcal{E})$, where each node $v_i\in \mathcal{V}$ represents a road link and each edge $e_{ij}\in \mathcal{E}$ represents the connectivity between adjacent links $v_i$ and $v_j$. 
    At time step $t$, the dynamic traffic features across the entire network are denoted as $\mathbf{X}^t \in \mathbb{R}^{N \times C}$, where $N:=|\mathcal{V}|$ and $C$ is the number of dynamic traffic features. 
    We denote $\mathbf{X}_i^t \in \mathbb{R}^{C}$ as the feature vector for link $v_i$ at time step $t$.
\end{myDef}

\begin{myDef}[Traffic Slice]
    A traffic slice for link $v_i$ at time step $\tau$ is defined as a set of dynamic traffic features $\mathbf{X}_i^{\tau-T+1:\tau} = [\mathbf{X}_i^{\tau-T+1}, \dots, \mathbf{X}_i^{\tau}] \in \mathbb{R}^{T \times C}$, where $T$ is the lookback window size. 
\end{myDef}

\begin{myDef}[Route]
    A route $\mathcal{R}$ is represented as a consecutive road link sequence
$\{v_i: i=1,\dots,l, v_i\in\mathcal{V}\}$, where $l$ is the total number of road links in route $\mathcal{R}$. 
Typically, a route consists of dozens to hundreds of links, and multiple trips can traverse the same route.
\end{myDef}


Then we formally define the focused problem in the paper.

\begin{prob}[Travel Time Estimation]
Given a trip query $q=(l_{ori}, l_{des}, \tau, \mathcal{R})$ starting at time step $\tau$, we aim to estimate the travel time $y$ from the origin $l_{ori}$ to the destination $l_{des}$ along the route $\mathcal{R}$ based on $\hat{y} = \mathcal{F}(q,\mathcal{G}, \mathbf{X}^{\tau-T+1:\tau}),$ where $\mathcal{F}(\cdot)=\mathcal{F}_r \circ \mathcal{F}_l(\cdot)$ is a mixture of route-centric and link-centric mapping functions we aim to learn based on historical queries $\mathcal{Q}=\{q_1,q_2,...\}.$
\end{prob}


\subsection{TTE System at DiDi} \label{subsec:didi_system}

As one of the leading ride-hailing platforms, DiDi has been continuously improving its TTE system. In this section, we brief the current TTE system at DiDi to facilitate further discussion.




\subsubsection{Route-centric backbone} 
\label{subsubsec:wdr}
DiDi's TTE system is built upon a route-centric Wide-Deep-Recurrent (WDR)~\cite{wdr2018kdd} architecture to learn intricate route-level feature interactions and structural dependencies. 
The \textit{wide} network memorizes low-level feature interactions by dedicated feature engineering followed by a generalized linear model. 
In contrast, the \textit{deep} network adopts Multi-Layer Perceptrons (MLPs) to learn high-order nonlinear feature interactions for improving generalizability. 
The \textit{recurrent} network models sequential dependencies among the road links of a route. 
The final travel time prediction is obtained by aggregating the outputs of the three modules.
Despite being introduced in 2018, WDR remains central to DiDi's TTE system as it enables modular maintenance and flexible upgrade. 
Leveraging the platform's abundant dynamic and static features, the wide module has undergone continual feature engineering refinement, while the recurrent module has been upgraded to a more advanced Transformer architecture~\cite{ieta2023kdd}.
\eat{
Grounded in the DiDi platform, abundant dynamic and static features can be extracted from vast ride-hailing datasets. Thoughtful consideration of the practical meanings and physical correlations of these features enables their effective memorization, thus providing a solid foundation for training deep neural networks while also ensuring interpretability—a crucial aspect for industrial applications. Furthermore, WDR's architecture is highly scalable; state-of-the-art deep learning models can seamlessly replace its deep or recurrent modules to further enhance performance. For instance, DiDi has already upgraded the original LSTM-based recurrent module to the advanced Transformer architecture.
}

\subsubsection{Incremental update}
\label{subsubsec:ieta}
In spite of the highly optimized model architecture, DiDi has also deployed an Incremental Learning~(IL) framework, iETA, to cost-effectively adapt models to constantly evolving traffic environments on a daily basis~\cite{ieta2023kdd}. 
Formally, let $\mathcal{D}_r$ denote the set of data collected on day $r.$ 
The model that serves on day $s$ is trained on the dataset $\mathcal{D}_{\mathrm{day}} = \mathcal{D}_{s-\delta_d} \cup \mathcal{D}_{s-7} \cup \cdots \cup \mathcal{D}_{s-7F},$ where $\delta_d<7$ indicates the delay for data preprocessing, and $F$ is the number of previous weeks considered for periodic data replay.
Without full retraining, iETA maintains the model up-to-date while consolidating recurrent traffic patterns for stability. 
This framework also supports various update frequencies to accommodate different scales of traffic distribution shifts.

%% file: method.tex
\section{Methodology}
\begin{figure*}[t]
    \centering
    \includegraphics[width=0.98\linewidth]{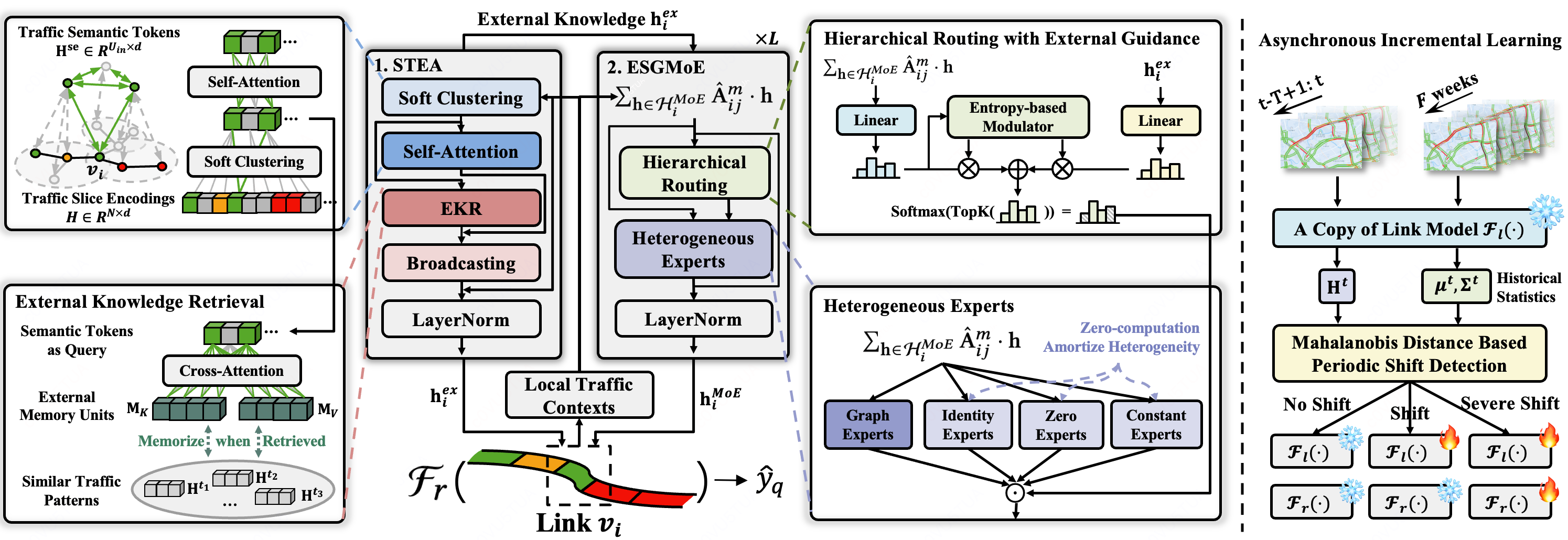}
    \caption{Overall framework of \model. The left part illustrates the pipeline of multi-level TTE from the view of a specific road link. The right part depicts the asynchronous strategy at the start of each IL update step.} 
    \label{fig:overall_framework}
\end{figure*}

As depicted in \figref{fig:overall_framework}, we tackle two major tasks for establishing \model: i) designing an expressive and scalable link representation learning module to enhance the route-centric model with rich traffic contexts, and ii) tailoring an optimization strategy for the mixture model to efficiently adapt to the changing traffic conditions at high frequency.
For the first task, we propose a Spatio-Temporal External Attention~(STEA) module to capture global spatio-temporal correlations, followed by Externally Stabilized Graph Mixture-of-Experts (ESGMoE) layers to learn more fine-grained but heterogeneous traffic patterns.
The link representations output by these two modules are then concatenated with the original in-route link features, serving as an enriched input for the downstream route-centric model.
For the second task, we propose Asynchronous Incremental Learning (ASIL) to efficiently and adaptively update the mixture model based on detected distribution shifts.

\subsection{Spatio-temporal External Attention} \label{subsec:stea}
As aforementioned, real-world urban road networks feature millions of road links, making scalable global correlation modeling a significant challenge.
Inspired by recent advancements~\cite{eamlp2023tpami,gea2024icml}, we propose a Spatio-Temporal External Attention~(STEA) module that learns a small set of external memory units $\mathbf{M} = [\mathbf{M}_1,\dots,\mathbf{M}_{U_{ex}}] \in \mathbb{R}^{U_{ex} \times d},$ where $U_{ex} \ll N$ and $d$ is the embedding dimension, to mediate the costly pairwise feature interaction among traffic slices.




Specifically, for traffic slices $\mathbf{X}^{t-T+1:t},$ we first leverage a lightweight MLP to encode temporal dependencies into high-dimensional representations $\mathbf{H} \in \mathbb{R}^{N \times d}$.
Then, we conduct External Knowledge Retrieval (EKR) from the memory units $\mathbf{M}$ via a cross-attention mechanism to augment the distinctiveness of $\mathbf{H}$, \ie
\begin{align}
    \begin{split}
    \operatorname{EKR}\left(\mathbf{H}, \mathbf{M}\right)  &= \operatorname{Cross-Attention}(\mathbf{H}, \mathbf{M}) \\
    &= \text{Softmax}\left((\mathbf{H}\mathbf{W}_Q)(\mathbf{M}\mathbf{W}_K)^\top / \tau\right) (\mathbf{M}\mathbf{W}_V)
    \end{split}
\end{align}
where $\mathbf{W}_Q, \mathbf{W}_K, \mathbf{W}_V\in \mathbb{R}^{d \times d}$ are learnable linear transformations, and $\tau$ is a temperature hyperparameter. 
As an alternative, we can directly replace $\mathbf{M}\mathbf{W}_K, \mathbf{M}\mathbf{W}_V$ with two distinct learnable external memory units $\mathbf{M}_K, \mathbf{M}_V \in \mathbb{R}^{U_{ex} \times d}$ to escalate model capacity.
Finally, the overall STEA mechanism can be formalized as:
\begin{align}
    \begin{split}
    \mathbf{H}^{ex} &= \operatorname{STEA}\left(\mathbf{H}, \mathbf{M}\right) = \operatorname{LN}\left( \operatorname{EKR}(\mathbf{H}, \mathbf{M}) + \mathbf{H} \right),
    \end{split}
\end{align}
where $\operatorname{LN}(\cdot)$ denotes layer normalization operator.
Through end-to-end training, the memory units are continuously updated by gradients from each traffic slice, enabling them to gradually accumulate prototypical traffic patterns across the entire dataset.
As a result, each EKR step allows a traffic slice to indirectly interact with all traffic slices across space and time. Similar traffic patterns will retrieve similar knowledge, leading to more concentrated representations in the latent space.
Notably, the EKR step has linear complexity $\mathcal{O}(N \cdot U_{ex}),$ not only achieving more scalable intra-time global modeling compared to quadratic complexity of pairwise modeling methods, but also filling the gap of coalescing temporally distant traffic slices.

In practice, intra-time traffic slices may exhibit more prominent semantic correlations, such as city-scale extreme congestion. 
To this end, we seamlessly integrate EKR with spatial hierarchical modeling~\cite{sshgnn2022tkde,transolver2024icml} to reinforce intra-time global correlation modeling.
We first conduct soft clustering on $\mathbf{H}$ to obtain traffic semantic tokens $\mathbf{H}^{se} \in \mathbb{R}^{U_{in} \times d}$, where $U_{in} \ll N.$ 
Then, we perform the EKR step at the semantic level to augment the intra-time Self-Attention~(SA) among semantic tokens. 
Finally, we BroadCast~(BC) back $\mathbf{H}^{se,ex}$ to yield link representations $\mathbf{H}^{ex}$ according to the soft clustering weights.
The above process can be formalized as
\begin{align}
\begin{split}
    \mathbf{H}^{ex} &= \operatorname{STEA}(\mathbf{H},\mathbf{M}) \\
    &= \operatorname{LN}\left( \operatorname{BC}\left( \operatorname{Self-Attention}\left(\mathbf{H}^{se}\right) + \operatorname{EKR}\left(\mathbf{H}^{se}, \mathbf{M}\right)\right) + \mathbf{H} \right).
\end{split}
\end{align} 
 We provide more details on hierarchical modeling in~\appref{app:intra_time}.

\subsection{Externally Stabilized Graph MoE}



Beyond STEA's global modeling, we further propose an Externally Stabilized Graph Mixture-of-Experts~(ESGMoE) layer to efficiently capture heterogeneous local traffic patterns with long-tailed distributions. 
In this section, we first provide an overview of the ESGMoE layer, followed by details on the routing and expert designs. 

\subsubsection{Overview of ESGMoE layer}
Overall, ESGMoE leverages a pool of $N_e$ graph experts to capture the full spectrum of heterogeneous traffic patterns while keeping low inference costs via sparse activation.
Formally, an ESGMoE layer is defined as
\begin{align}
    {\mathbf{h}_i^{MoE}}^{'} &= LN\left( \sum_{n=1}^{N_e} G_n \left( \mathcal{H}_i^{MoE}, \mathbf{h}_i^{ex} \right) \cdot E_n\left(\mathcal{H}_i^{MoE} \right) + \mathbf{h}_i^{MoE} \right), 
\end{align}
\eat{
where it encodes local traffic contexts within each link's $m$-hop neighborhood $\mathcal{N}_m(i)$, \ie $\mathcal{H}_i^{MoE} = \{\mathbf{h}_j^{MoE} : j \in \mathcal{N}_m(i) \cup \{i\}\},$ into compact representations ${\mathbf{h}_i^{MoE}}^{'} \in \mathbb{R}^{d}.$}
which encodes local traffic contexts  $\mathcal{H}_i^{MoE} = \{\mathbf{h}_j^{MoE} : j \in \mathcal{N}_m(i) \cup \{i\}\}$ within each link's $m$-hop neighborhood $\mathcal{N}_m(i)$ into a compact representation ${\mathbf{h}_i^{MoE}}^{'} \in \mathbb{R}^{d}.$
Here, $\mathbf{h}_j^{MoE}$ denotes the initial traffic features, which is the same as the inputs of the STEA module, or the hidden representation output by the previous ESGMoE layer for link $v_j$, which preserves critical temporal patterns while avoiding costly graph modeling for each time step.
The gate function $\mathbf{G}(\cdot)$ and graph expert $E_n(\cdot)$ lie in the core of its design.
Specifically, $G_n(\cdot)$ is the $n$-th element of a sparse gate function $\mathbf{G}(\cdot)$ that determines the probability of routing link $v_i$ to the $n$-th expert $E_n(\cdot)$ based on the local traffic contexts and external knowledge $\mathbf{h}_i^{ex}$ retrieved from the STEA module.
The expert designs are heterogeneous.
Most experts are implemented as a simplified graph convolution network that has low computational overhead~\cite{sgc2019icml}, \ie $E_n(\mathcal{H}_i^{MoE}) = \operatorname{MLP}_n(\sum_{\mathbf{h} \in \mathcal{H}_i^{MoE}} \hat{\mathbf{A}}^m_{ij} \cdot \mathbf{h})$, 
where $\hat{\mathbf{A}}^m$ is the $m$-th power of the normalized adjacency matrix $\hat{\mathbf{A}}$ that allows pre-computation for inference acceleration.
The remaining experts are zero-computation experts designed to amortize over expert activations for common traffic patterns. 




\subsubsection{Hierarchical routing with adaptive external guidance}
\label{subsubsec:hier_route}
A well-established routing mechanism in MoE is essential for enabling stable expert specialization~\cite{moesurvey2022}. 
Existing spatio-temporal MoE methods typically feed into the routing gate the short-term traffic slices~\cite{stmoe2023cikm,testam2024iclr} or learnable link and time embeddings that capture long-term patterns~\cite{cpmoe2024kdd}, which either lack discriminative power or limit the gate to learning fine-grained short-term distinctions.
To this end, we propose an entropy-based hierarchical routing mechanism to adaptively enhance the distinctiveness of gate inputs with the external knowledge accumulated by the STEA module. 

Specifically, the routing mechanism adopts the classical Top-$k$ gate function~\cite{switch2022jmlr} to produce the routing probabilities for graph experts, \ie
$\mathbf{G}(\cdot) = \operatorname{Softmax} (\operatorname{Topk}(\mathbf{g}(\cdot))),$
where $\mathbf{g}(\cdot)$ is usually implemented as a linear layer that maps the input features to routing logits, and $k$ is the number of activated experts for each sample.
A load balancing loss is also equipped to prevent the sparse gate from collapsing to a few dominated experts, which will be introduced in~\secref{subsec:objs}.
Differently, we design $\mathbf{g}(\cdot)$ to possess a hierarchical structure as follows:
\begin{align}
\mathbf{g}(\mathcal{H}_i^{MoE}, \mathbf{h}_i^{ex}) = \alpha_i \cdot \mathbf{g}^{ex}(\mathbf{h}_i^{ex}) + (1 - \alpha_i) \cdot \mathbf{g}^{loc}(\mathcal{H}_i^{MoE}),
\end{align}
where $\mathbf{g}^{ex}(\cdot)$ and $\mathbf{g}^{loc}(\cdot)$ are two separate FFNs that compute the routing logits of link $v_i$ using externally enhanced link representation and local traffic contexts, respectively. 
The adaptive weight $\alpha_i$ is derived from the entropy $S_i$ of the local routing distribution:
\begin{align}
\begin{split}
\alpha_i &= \text{Sigmoid}\left( \gamma  S_i + \mu \right), \\
S_i = - \mathbf{\overline{g}}^{loc}_{i} \cdot \log &\mathbf{\overline{g}}^{loc}_{i}, \quad
\mathbf{\overline{g}}^{loc}_{i} = \text{Softmax}\left(\mathbf{g}^{loc}(\mathcal{H}_i^{MoE}) \right)
\end{split}
\end{align}
where $\gamma, \mu$ are learnable parameters.
In this way, when the router is uncertain about local traffic contexts, $\alpha_i$ will increase to put more reliance on external guidance for routing stability, which enables more robust expert specialization.

\subsubsection{Amortizing heterogeneity with zero-computation experts}
\eat{
As shown in \figref{fig:imbalance}, the heterogeneous traffic patterns in real-world road networks are in fact skewedly distributed in multiple dimensions. 
To fully capture such diverse patterns, vanilla MoE designs fall short from two perspectives:\TODO{too long?}
i) To facilitate the expert specialization on rare traffic patterns, those experts who should specialize in rare traffic patterns should not be activated too frequently on common patterns, which contradicts the load balancing regularization of MoE. Simply increasing the total number of experts can amortize the expert load, but may lead to excessive model capacity and the undertraining of certain experts.
ii) For those underrepresented traffic patterns, more experts should be activated to strengthen the pattern learning and produce robust representations with proper expert ensembling. However, the sparsity-driven top-$k$ activation strategy is not flexible enough to adaptively adjust the number of activated experts. Simply increasing the activation number will lead to excessive computational costs and may even cause overfitting on common patterns.
}
Despite the potential of training specialized experts to capture heterogeneous traffic patterns, vanilla sparse-gated MoE designs are inherently limited when facing the long-tailed patterns:
(i) Experts meant for rare patterns are often activated too much on common patterns due to load balancing regularization, and simply increasing expert numbers can cause overcapacity and undertraining; (ii) Rare patterns should be assigned with more experts, but top-$k$ sparse activation lacks flexibility. Naively raising $k$ will increase computation and risk overfitting common patterns.
To this end, we further introduce zero-computation experts that do not perform graph modeling to amortize the activations of common patterns and implicitly allow more experts to be activated for rare patterns.

Inspired by work~\cite{moeplus2025iclr}, we introduce three types of zero-computation experts: i) \textit{Identity expert} that simply returns the input feature; ii) \textit{Constant expert} that outputs a learnable constant vector; iii) \textit{Null expert} that outputs a zero vector.
The activation of these experts is determined simultaneously with the graph experts by the same routing mechanism introduced in~\secref{subsubsec:hier_route}.
Intuitively, 
the three types of zero-computation experts have their own strengths in dealing with certain types of common traffic patterns: i) identity experts for smooth/free-flow or near-linear regimes where the current representation is already predictive; ii) constant experts to memorize highly frequent, template-like patterns via a learned bias; iii) null experts to bypass graph modeling when signals are noisy or the marginal gain is negligible. Therefore, introducing these experts can effectively amortize frequent, low-marginal-gain traffic patterns without overly consuming graph-expert capacity.


\subsection{Asynchronous Incremental Learning}
Continuous adaptation is crucial for industrial TTE systems to maintain stability in evolving traffic contexts. 
However, the introduction of the link-centric model renders frequent full-parameter updates computationally costly and more susceptible to overfitting diverse link-level distribution shifts.
To this end, we propose an ASynchronous Incremental Learning~(ASIL) strategy that actively detects distribution shifts to enable selective parameter updates.

First, since periodic traffic patterns are extensively learned by the model, we invoke the update only when periodic distribution shifts are detected.
Specifically, we maintain historical statistics of representations for each link $v_i$, including mean $\boldsymbol{\mu}_i^t$ and covariance $\boldsymbol{\Sigma}_i^t$, which are computed using link representations from the same time step over the past $F$ weeks.
At the start of an update, for each link, we compute the Mahalanobis Distance~(MD) between its current representation $\mathbf{h}_i^t$ and the historical statistics to quantify the periodic shift degree:
\begin{align}
d_i^t = \sqrt{(\mathbf{h}_i^t - \boldsymbol{\mu}_i^t)^\top (\boldsymbol{\Sigma}_i^t)^{-1} (\mathbf{h}_i^t - \boldsymbol{\mu}_i^t)}.
\end{align}
A link with $d_i^t$ exceeding a preset threshold $\delta_d$ is considered anomalous, \ie it is experiencing a periodic shift.
Only when the proportion of anomalous links across the city exceeds the $\delta_l$~quantile of its historical distribution, do we invoke the IL update step.
Moreover, considering that the route-centric model naturally focuses more on the relatively stable structural dependencies when fed with link-level contexts, we impose stricter restrictions on route-side parameter updates. 
Only if the anomaly proportion surpasses the $\delta_r$~quantile of its historical distribution will we trigger the link-side and route-side updates simultaneously, where $\delta_r > \delta_l$.

However, as the link encoder is incrementally updated, the latent space for link representations changes across model versions, which may affect the reliability of MD. To address this, we use a frozen copy of the link-centric model from the daily update to generate representations for drift detection, ensuring consistent latent space with low extra inference cost.

\subsection{Optimization Objectives}
\label{subsec:objs}
The optimization objectives of \model consist of two parts.
The first part is the TTE regression loss that jointly optimizes the route- and link-centric models in an end-to-end manner, \ie
\begin{align}
    \mathcal{L}_{reg} = \sum_{q\in \mathcal{Q}} | \hat{y}_q - y_q |.
\end{align}
The second part is an expert load balancing regularizer that prevents ESGMoE layers from model collapsing issues, \ie
\begin{align}
    \mathcal{L}_{load} = N_e \cdot \sum_{n=1}^{N_e} \xi_n \cdot \left( \frac{1}{N} \sum_{i=1}^N G_{n,i} \right) \cdot \left( \frac{C_n}{N \cdot k} \right)
\end{align}
where $C_n$ is the count of tokens routed to expert $n$ under top-$k$ selection. $\xi_n$ is the expert type weight which is set to $1$ for graph experts and $\delta$ for zero-computation experts, where $\delta$ is a hyperparameter that controls the extent of amortization.

Overall, we train \model by jointly optimizing the objectives
$
    \mathcal{L} = \mathcal{L}_{reg} + \alpha \sum_{l=1}^L\mathcal{L}_{load}^{(l)},
$
where $\mathcal{L}_{load}^{(l)}$ is the load balancing loss of the $l$-th ESGMoE layer,
$\alpha$ is a hyperparameter that controls the extent of expert balancing.

%% file: deployment.tex
\section{System Deployment}\label{sec:deploy}


\model has been deployed in DiDi's ride-hailing platform to upgrade the existing route-centric TTE system. This section provides an overview of the system deployment strategies.

\subsection{Offline Model Training} \label{subsec:offtrain}
Offline model training for each city is conducted using TensorFlow on a Linux server with 4 Intel Xeon E5-2630 v4 CPUs~(90 GB) and 1 NVIDIA Tesla P40 GPU~(24GB). 
The model is incrementally updated under a two-level temporal hierarchy.
At the daily scale, the update follows a similar pipeline as iETA~\cite{ieta2023kdd}.
At the hourly scale, the ASIL strategy is employed based on the completed trip data recorded in the past hour. 
Offline training adopts time-specific batch sampling, where each batch comprises historical trip records originating at the identical time step $t$ rather than randomly sampled from the entire dataset.
This yields three key benefits: (1) it leverages global modeling within the same periods to ensure smoother gradients for link representations, (2) it improves training efficiency, as trips sharing the same road links at a specific time step can reuse the computed link representations during forward and backward propagation, and (3) it aligns with the streaming inference paradigm in the online environment. 
Upon completion, the model will be released on online servers to enable real-time TTE services.

\subsection{Asynchronous Online Serving}
\eat{
The online inference system follows a Client-Server~(C/S) architecture. 
End-users interact with client applications, which communicate via API calls with backend servers.}
On the online server side, to meet the high throughput and ms-level latency requirements of real-time response, we adopt an asynchronous online serving strategy for \model.
The link-centric model is implemented in TensorFlow and deployed on GPU servers, leveraging parallel expert inference for large-scale link representation generation. Due to the higher computational cost, link representations are updated per 5 minutes and cached in a Redis server.
The route-centric model is re-implemented in C++ and deployed on distributed CPU servers. 
When a TTE query arrives, the route-centric model retrieves the latest link representations along with route features for instant prediction.

%% file: experiments.tex
\section{Experiments}
In this section, we conduct extensive experiments to evaluate our proposed framework based on DiDi’s current TTE system.
In particular, we focus on:
(1) the overall performance of \model compared to the state-of-the-art TTE methods, 
(2) the long-tail performance of \model,
(3) the efficiency of \model,
(4) the ablation study of key modules in \model,
(5) the interpretability of \model, 
and (6) the efficacy of \model in online production environments.
Please refer to \appref{app:experiment} for more experimental details.

\subsection{Experiment Setup}

\subsubsection{Training preparation}
We conduct extensive experiments on three metropolises, Beijing, Nanjing and Suzhou.
All datasets are collected from DiDi's ride-hailing platform. 
Detailed statistics of three datasets are reported in \tabref{tab:stats}. 
It is worth noting that for each city, we identify a set of "hot links" with non-trivial traffic dynamics for link embedding generation, which helps reduce computation and link-level noise. See~\appref{app:hotlink} for details on hot link selection.
For each dataset, we take the last 7 days for testing. 
In the \textbf{full retraining} setting, we use the 110 days preceding each test day as the training set.
In the \textbf{incremental learning} setting, we take the trips completed in the last hour as the training set and the trips started in the next hour as the test set. This high-frequency setting is realistic and more suitable for testing the efficacy and stability of IL strategies.
The performances in all hours are averaged to obtain the overall results.
\begin{table}[t]
\caption{Statistics of three datasets.}
\begin{tabular}{@{}c|ccc@{}} 
\toprule
City     & Beijing & Nanjing & Suzhou \\ \midrule
Time Span    & \makecell{2024-07-21$\sim$\\2024-09-05} & \makecell{2024-12-18$\sim$\\2025-02-25} & \makecell{2025-01-01$\sim$\\2025-04-14} \\
\midrule
\# of Trips   & 22,276,096 & 7,954,432 & 16,775,680 \\
Avg. Duration (min) &  18.11   &  13.79   &  13.63  \\
\makecell[c]{Avg. \# of \\ In-route Links} &  121.01   &  99.92   & 86.86 \\ 
\midrule
\# of Links & 2,315,360 & 696,107 & 1,419,074 \\
\# of Hot Links & 156,731 & 83,323 & 149,171 \\
\bottomrule
\end{tabular}
\label{tab:stats}
\end{table}

\subsubsection{Baselines}


In the full retraining setting, we compare \model without IL~(\model-\textit{WoIL}) with the following baselines:
\textit{Rule-based:} 
(1) RouteETA: It predicts TTE by adding up the historical average travel time of in-route links.
\textit{Route-centric:} 
(2) HierETA~\cite{hiereta2022kdd}: A two-level hierarchical expert system for route-level TTE;
(3) WDR: The current TTE model deployed at DiDi without incremental learning, as introduced in~\secref{subsubsec:wdr};
\textit{Link-centric:} 
(4) CompactETA~\cite{compacteta2020kdd}: It uses graph attention networks to produce link representations, followed by lightweight MLPs for route TTE;
(5) ConSTGAT~\cite{constgat2020kdd}: It designed a 3D GAT for link TTEs, which are added up to obtain route TTE;
(6) BigST~\cite{bigst2024vldb}: It proposes a linearized global spatial modeling method, which is used to replace our link-centric model.

In the IL setting, we compare \model with \textit{route-centric} method iETA~\cite{ieta2023kdd}.
For fairness, all the methods are input with our curated features, and link-centric models are trained with the same pipeline as introduced in~\secref{subsec:offtrain}.
Please see \appref{app:imp_detail} for more implementation details.

\subsubsection{Metrics}
We use three metrics to evaluate the performance of different methods:
\textbf{Mean Absolute Error (MAE)}: the average absolute difference between the predicted and actual TTE;
\textbf{Mean Absolute Percentage Error (MAPE)}: the average absolute percentage difference between the predicted and actual TTE;
\textbf{Bad Case Ratio (BCR)}: the percentage of trip queries whose predicted TTE satisfies MAE $>$ 300 seconds and MAPE $>$ 20\%.

\subsection{Overall Performance (RQ 1)}
\tabref{tab:overall_performance} presents the overall performance. Our proposed method, \model, consistently outperforms all baselines across all metrics.
In the full retraining setting, \model achieves up to 2.01\%, 2.41\%, and 8.81\% relative improvements in MAE, MAPE, and BCR over the second-best result.
In the incremental learning (IL) setting, it achieves up to 2.39\%, 3.70\%, and 10.32\% relative improvements over iETA.
These gains demonstrate \model's effectiveness and adaptiveness in capturing multi-level, evolving traffic patterns. The more substantial BCR improvements highlight the efficacy of our link-level designs for long-tail generalization.
We can further make the following observations:
(1) Deep learning methods surpass the rule-based baseline (RouteETA), indicating their superiority in capturing high-order and nonlinear dependencies for accurate TTE;
(2) Link-centric baselines mostly underperform the industrial route-centric system (WDR), underscoring the importance of feature engineering in industrial practice;
(3) Integrating BigST outperforms WDR \wrt MAPE and BCR in Beijing, justifying the importance of link-level modeling. However, it fails to uniformly outperform WDR, indicating the nontriviality of integrating link-level modeling into highly optimized route-centric TTE systems;
(4) IL methods consistently outperform full retraining methods, showing IL's effectiveness for non-stationary traffic. However, the improvement of iETA over WDR is marginal, which is because of the lack of modularity in iETA's IL strategy.

\begin{table*}[thbp]
\centering
\caption{Overall performance comparison of different methods on three real-world datasets. The best results are in \textbf{bold}, the second best are \underline{underlined}, and the third best begin with a starisk ($^*$).}
\label{tab:performance}
\begin{tabular}{@{}c|c|ccc|ccc|ccc@{}}
\toprule
\multirow{2}{*}{\textbf{\makecell[c]{Training \\ Setting}}} & \multirow{2}{*}{\textbf{Model}} & \multicolumn{3}{c|}{\textbf{Beijing}} & \multicolumn{3}{c|}{\textbf{Nanjing}} & \multicolumn{3}{c}{\textbf{Suzhou}} \\
\cmidrule(l){3-11}
 & & MAE(sec) & MAPE(\%) & BCR(\%) & MAE(sec) & MAPE(\%) & BCR(\%) & MAE(sec) & MAPE(\%) & BCR(\%) \\
\midrule
No Training & RouteETA & 181.06 & 17.12 & 11.87 & 153.06 & 22.56 & 11.4 & 139.07 & 19.84 & 8.61 \\

\midrule 
\multirow{5}{*}{Full Retraining} & HierETA & 135.07 & 12.84 & 6.20 & 87.02 & 12.61 & 2.31 & 82.53 & 11.26 & 1.90 \\
 & WDR & 134.02 & 12.06 & 6.06 & 77.29 & 10.59 & 1.59 & 78.60 & 10.64 & 1.53 \\
 & CompactETA & 136.63 & 12.25 & 6.12 & 83.54 & 11.79 & 2.10 & 81.38 & 10.87 & $^*$1.46 \\
 & ConSTGAT & 134.52 & 12.11 & 6.04 & 78.16 & 10.56 & 1.98 & 79.82 & 10.73 & 1.59 \\
 & BigST & 134.75 & 11.98 & 6.03 & 77.93 & $^*$10.49 & 1.91 & 79.21 & 10.77 & 1.54 \\
 & \model-\textit{WoIL} & \underline{132.43} & \underline{11.86} & $^*$5.73 & \underline{75.74} & \underline{10.26} & \underline{1.45} & \underline{77.29} & \underline{10.51} & \underline{1.42} \\
\midrule
\multirow{2}{*}{\makecell[c]{Incremental \\ Learning}} & iETA & $^*$132.73 & $^*$11.93 & \underline{5.71} & $^*$77.10 & 10.53 & $^*$1.55 & $^*$78.22 & $^*$10.59 & $^*$1.46 \\
 & \model & \textbf{130.84}	& \textbf{11.75} & \textbf{5.25}	& \textbf{75.26}	& \textbf{10.14}	& \textbf{1.39} & \textbf{77.18} & \textbf{10.48} & \textbf{1.36} \\
\bottomrule
\end{tabular}
\label{tab:overall_performance}
\end{table*}

\subsection{Long-tail Performance (RQ 2)}
\begin{figure}[t] 
\centering  
\subfigure[Trip duration.]{ 
    \includegraphics[width=0.31\linewidth]{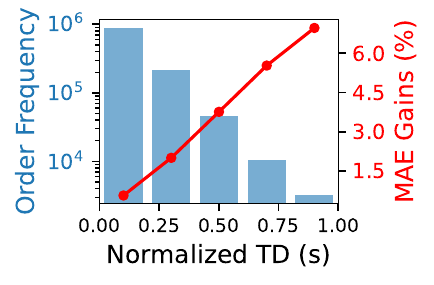}}
\subfigure[En-route traffic condition deviation degree.]{ 
    \includegraphics[width=0.31\linewidth]{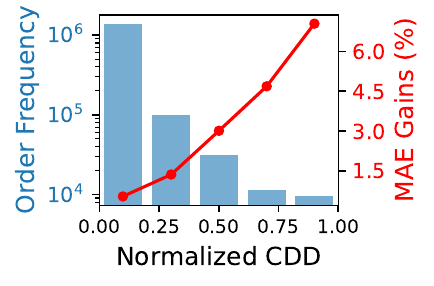}}
\subfigure[En-route traffic condition non-recurrence degree.]{
    \includegraphics[width=0.31\linewidth]{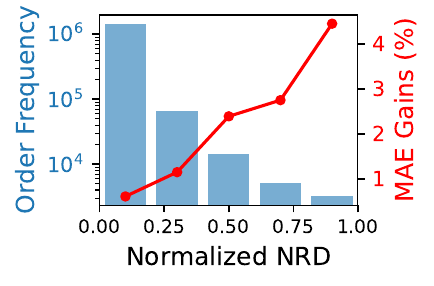}}
\caption{ 
Relative MAE gains of MixTTE over DiDi's current TTE model~\cite{ieta2023kdd} across head and tail traffic scenarios.
}
\label{fig:longtail_gains}
\end{figure}
We additionally analyzed the performance gains in long-tail traffic scenarios. 
Specifically, we first divided the test trip queries in Suzhou dataset into 5 groups by evenly partitioning the value range of the following three metrics:
(1) the Trip Duration (TD), 
(2) en-route traffic Condition Deviation Degree (CDD) 
and (3) en-route traffic condition Non-Recurrence Degree (NRD).
Here, the metrics CDD and NRD for trip query $q=(l_{ori}, l_{des}, \tau, \mathcal{R})$ are defined as
\begin{align*}
    \text{CDD}(q) = \mathbb{E}_{v \in \mathcal{R}} \|x_v^{\tau_v} - x_v^\tau \|_2, \quad
    \text{NRD}(q) = \mathbb{E}_{v \in \mathcal{R}} \|x_v^{\tau_v} - \overline{x}_v^{\tau_v} \|_2,
\end{align*}
where $x_v^\tau$ denotes the discrete traffic congestion level feature for link $v$ at time step $\tau.$ Time step $\tau_v$ is when the vehicle arrives at link $v$, and $\overline{x}_v^{\tau_v}$ denotes the historical average congestion level \wrt time step $\tau_v.$
As shown in \figref{fig:longtail_gains}, the ride-hailing trips exhibit long-tail distributions \wrt all three metrics.
More importantly, after calculating the MAE gains of MixTTE over DiDi's current TTE model in different sample groups, we can observe that the MAE gains exhibit an increasing trend from head to tail samples, which validates MixTTE's generalizability on long-tail cases.

\subsection{Efficiency Analysis (RQ 3)}
We further conduct experiments to verify the efficiency of \model.
In the full retraining setting, we compare \model-\textit{WoIL} to the link-centric baselines that share the linear complexity \wrt link number. 
We use the training throughput and inference latency as our metrics, where
the throughput is measured by the average number of training samples processed per minute and the inference latency is only measured for link representation generation at each time step.
For fair comparison, we align the baselines with \model in terms of the activated parameters per inference step and the neighborhood size for local traffic modeling.
As shown in \tabref{tab:efficiency1}, \model-\textit{WoIL} consistently outperforms CompactETA and ConSTGAT across all three efficiency metrics. This is because CompactETA needs to stack multiple costly GAT layers to capture long-range dependencies, while the 3D GAT adopted in ConSTGAT jointly performs spatial and temporal attention, making it $T$ times more expensive than a single GAT layer.
Besides, our model achieves comparable efficiency with BigST while possessing a larger model capacity, which justifies the efficacy of the ESGMoE module in increasing the model capacity without incurring extra computational costs. 

\begin{table}[t]
\caption{Full retraining and inference efficiency comparison. (BJ: Beijing, NJ: Nanjing, SZ: Suzhou).}
\begin{tabular}{@{}c|ccc|ccc@{}} 
\toprule
\multirow{2}{*}{Model} & \multicolumn{3}{c|}{\makecell[c]{Training Throughput \\ (K/min)}} & \multicolumn{3}{c}{\makecell[c]{Inference Latency \\ (msec/time step)}} \\
\cmidrule(l){2-7} 
& BJ & NJ & SZ & BJ & NJ & SZ \\
\midrule
CompactETA & 14.96 & 23.13 & 16.23 & 541.27 & 286.62 & 476.84 \\
ConSTGAT & 10.51 & 17.80 & 12.64 & 704.11 & 358.52 & 650.98 \\
BigST & \textbf{22.45} & \textbf{32.77} & \underline{23.53} & \textbf{78.30} & \underline{40.05} & \textbf{69.40} \\
\model-\textit{WoIL} & \underline{21.17} & \underline{32.02} & \textbf{25.38} & \underline{88.97} & \textbf{30.45} & \underline{75.11} \\
\bottomrule 
\end{tabular} 
\label{tab:efficiency1} 
\end{table}

In the IL setting, we compare our framework with ~\emph{-WoPU}, which removes the parameter-selective update in the ASIL strategy. We use average training time and trainable parameters over all IL steps as our metrics. We only report the results on the Suzhou dataset since the ASIL strategy is not sensitive to the link number.
As shown in \tabref{tab:efficiency2}, \model achieves a significant reduction in training time and trainable parameters compared to \model-\textit{FIL}. While \model-\textit{FIL} updates all the parameters at each IL step, \model selectively activates route- or link-level parameters only when periodic shifts occur, thereby achieving promising efficiency gains.
\begin{table}[t]
\caption{IL efficiency comparison in Suzhou dataset.}
\begin{tabular}{c|cc} 
\toprule
Model & \makecell[c]{\makecell[c]{Avg. Training Time \\ (sec/step)}} & \makecell[c]{Avg. Trainable Params \\ (K/step)} \\
\midrule 
\model-\textit{WoPU} & 36.63 & 525.66 \\
\model & 7.81 & 47.49 \\
\bottomrule
\end{tabular}
\label{tab:efficiency2}
\end{table}

\subsection{Ablation Study (RQ 4)}
\begin{figure}[t]
\centering  
\includegraphics[width=1\linewidth]{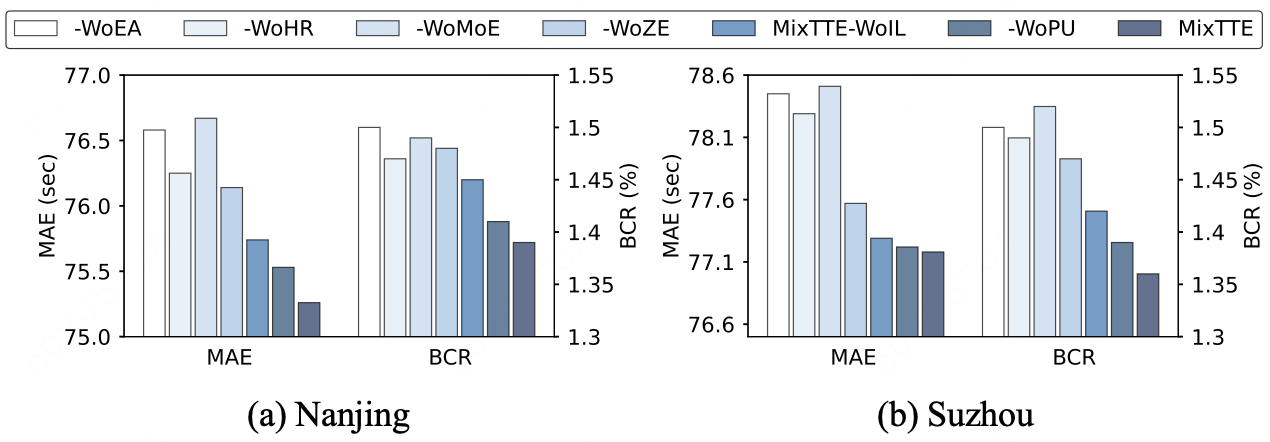}
\vspace{-5pt}
\caption{Ablation study on Nanjing and Suzhou datasets. }
\vspace{-5pt}
\label{fig:ablation}
\end{figure}

To justify the efficacy of each module in \model, we compare the following model variants on the Nanjing and Suzhou datasets:
i)~\emph{-WoEA} removes the STEA module along with the hierarchical routing in the ESGMoE layer;
ii)~\emph{-WoHR} removes the hierarchical routing in the ESGMoE layer;
iii)~\emph{-WoMoE} removes the ESGMoE layers;
iv)~\emph{-WoZE} removes the zero-computation experts in the ESGMoE layer;
v)~\emph{-WoPU} as defined in the efficiency analysis.
As shown in~\figref{fig:ablation}, we make the following observations.

In the full retraining setting, \emph{-WoHR} is worse than \model-\emph{WoIL}, which indicates its importance of hierarchical routing in stabilizing ESGMoE training.
Comparing \emph{-WoEA} with \emph{-WoHR}, we can see that STEA is crucial for enhancing route contextual awareness.
Moreover, \emph{-WoHR} causes a larger performance drop in the Suzhou dataset, which is because the more heterogeneous road network in Suzhou demands more stable MoE training.
Finally, the performance drop caused by \emph{-WoMoE} reveals the capacity of ESGMoE layers; \emph{-WoZE} also show performance drops, especially \wrt BCR in the Nanjing dataset, demonstrating the importance of zero-computation experts in enhancing long-tail scenarios.
In the IL setting, \emph{-WoPU} causes performance drops, indicating its efficacy in boosting the stability of IL for the established mixture TTE model. 






\subsection{Interpretability Analysis (RQ 5)}
We conduct additional interpretability analyses on the expert activation distributions of ESGMoE layers in different traffic scenarios to provide more insights into the model's decisions.
Specifically, at each time step, we compute the Top-2 routing distributions for every link and record the corresponding expert assignments. For each day, we obtain a link–expert assignment tensor $\mathbf{A} \in \mathbb{R}^{288 \times N \times 16}$, where $N$ denotes the number of links, $16$ is the number of experts~(the 0-11th are graph experts, the 12th is a null expert, the 13th is an identity expert, the 14-15th are constant experts), and 288 is the total number of 5-minute time steps in a day. 
This tensor enables us to analyze the traffic scenarios that each expert specializes in.
\begin{figure}[t] 
\centering  
\subfigure[Expert specialization in layer 0 in congested scenarios.]{ 
    \includegraphics[width=0.90\linewidth]{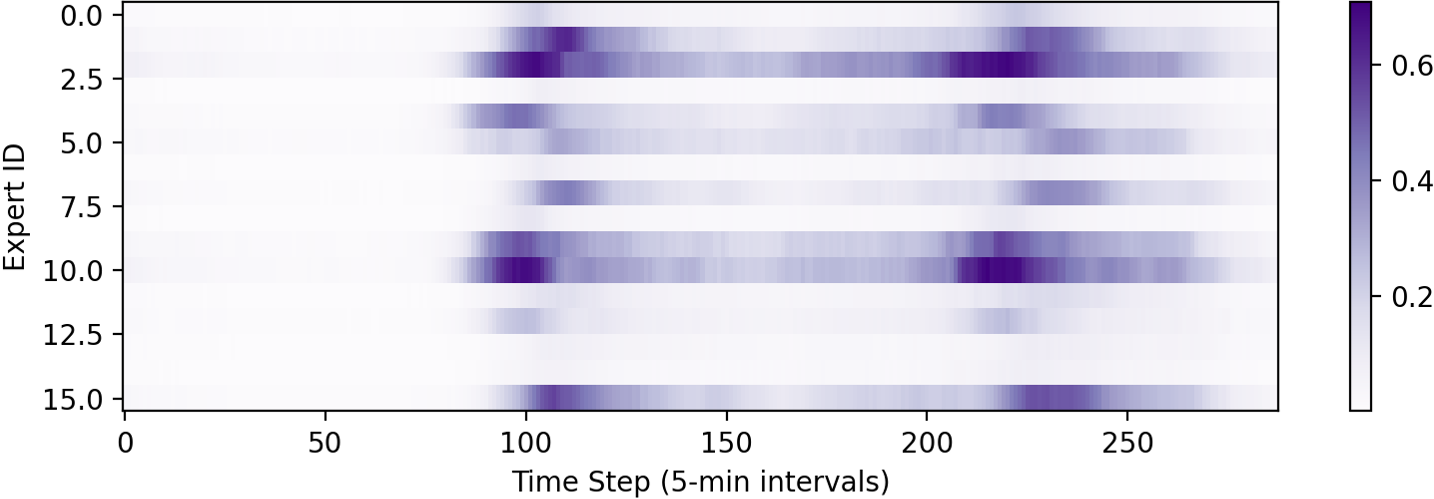}}
\subfigure[Expert specialization in layer 1 in congested scenarios.]{
    \includegraphics[width=0.90\linewidth]{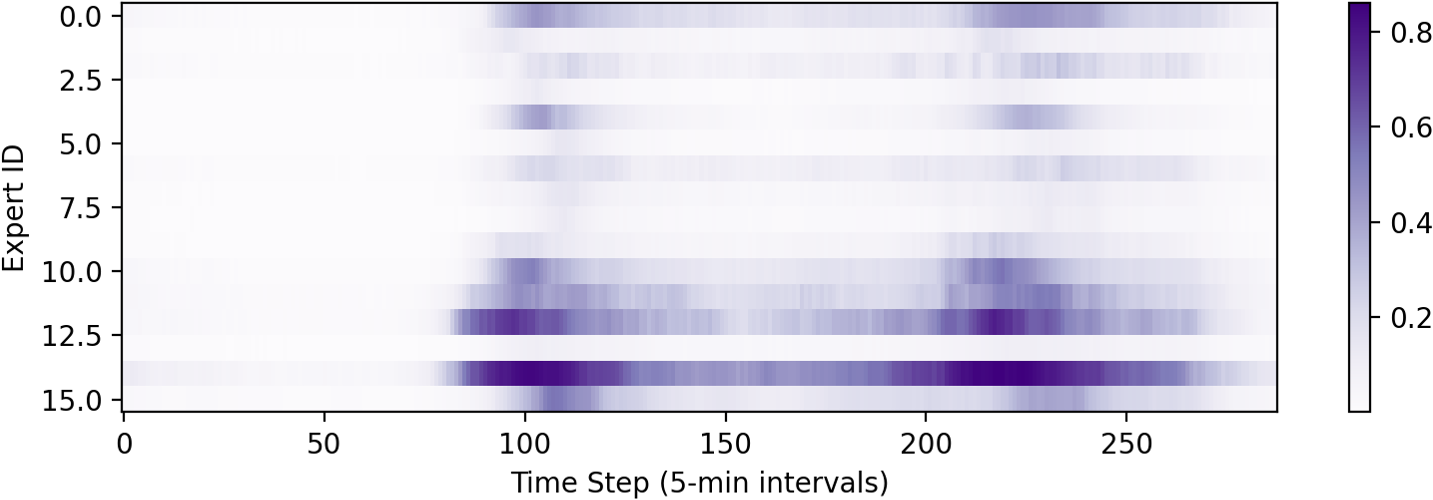}}
\subfigure[Expert specialization in layer 0 in non-recurring scenarios.]{ 
    \includegraphics[width=0.92\linewidth]{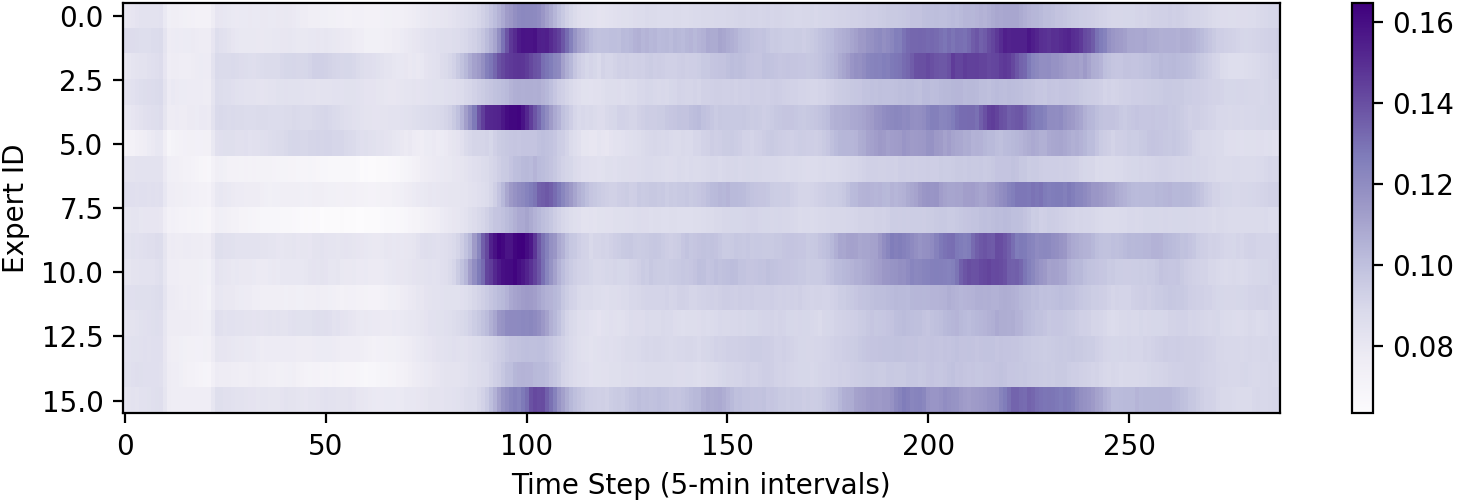}}
\subfigure[Expert specialization in layer 1 in non-recurring scenarios.]{
    \includegraphics[width=0.92\linewidth]{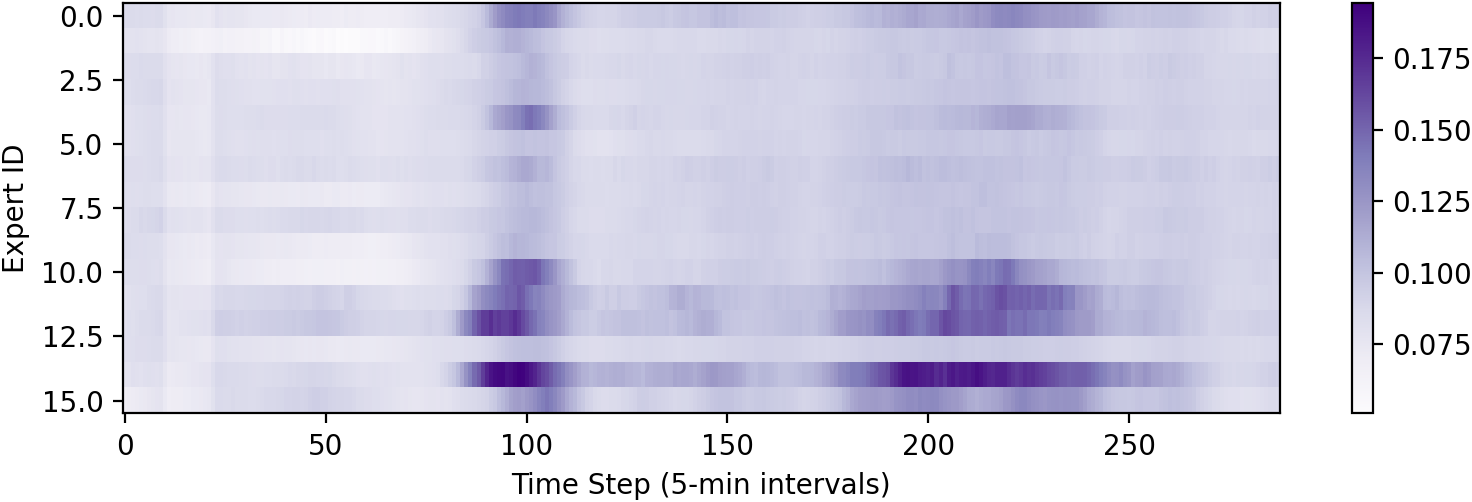}}
\caption{ 
Interpretability analysis of ESGMoE layers in Beijing dataset on 2024-09-02. The color of each matrix element indicates the proportion of links assigned to a given expert that are experiencing congestion or non-recurring conditions at each time step. 
Here, the condition is defined as non-recurrent if the deviation between the current traffic condition and its historical average exceeds the 0.9 quantile of the overall historical distribution.
}
\label{fig:interpret}
\end{figure}

As shown in \figref{fig:interpret}, we can obtain several key findings.
(1)~\textit{Expert specialization:} A subset of experts are activated for modeling congested or non-recurring traffic patterns, which indicates that our heterogeneous expert design can effectively form specialization towards long-tail scenarios. 
(2)~\textit{Layer-wise distinction:} Comparing the activation distributions in different layers, there is a sparser expert specialization in long-tail scenarios in layer 1 compared with that in layer 0, which is consistent with the fact that deeper layers tend to capture more abstract traffic semantics.
(3)~\textit{Behavior of zero-computation experts:} 
First, they are much less activated than graph experts in Layer 0. This is because raw feature modeling relies heavily on graph computation to capture high-level patterns. 
Second, their usage increases in Layer 1, where condensed semantics from the previous layer allow for more lightweight processing to avoid over-modeling. 
Third, constant experts show consistently high engagement in both layers, reflecting their capacity for memorization. Null and identity experts, while less frequent, exhibit meaningful activation patterns in non-recurring scenarios and higher layers, confirming their necessity in amortizing simple or noisy patterns.

\subsection{Online Evaluation (RQ 6)}
In this section, we conduct online A/B testing at DiDi's large-scale ride-hailing platform to validate \model's utility in real-time production environments. 
Particularly, we focus on TTE for drop-off trip queries at the beginning of each trip, which covers a wide range of trip distances and durations.
During the testing period, we randomly split the trip queries into a control group and a treatment group, where the control group is answered by the current TTE system at DiDi, and the treatment group is answered by the system enhanced with \model. 
Once a trip is completed, we record its actual travel time for further metric calculation.

\begin{figure}[t] 
\centering  
\subfigbottomskip=-3pt 
\subfigcapskip=-2pt 
\subfigure[City X.]{ 
    \includegraphics[width=0.45\linewidth]{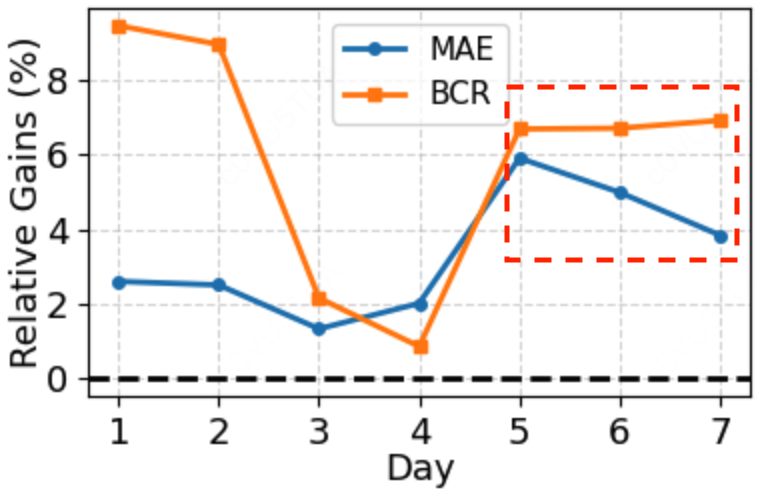}}
\quad
\subfigure[City Y.]{
    \includegraphics[width=0.47\linewidth]{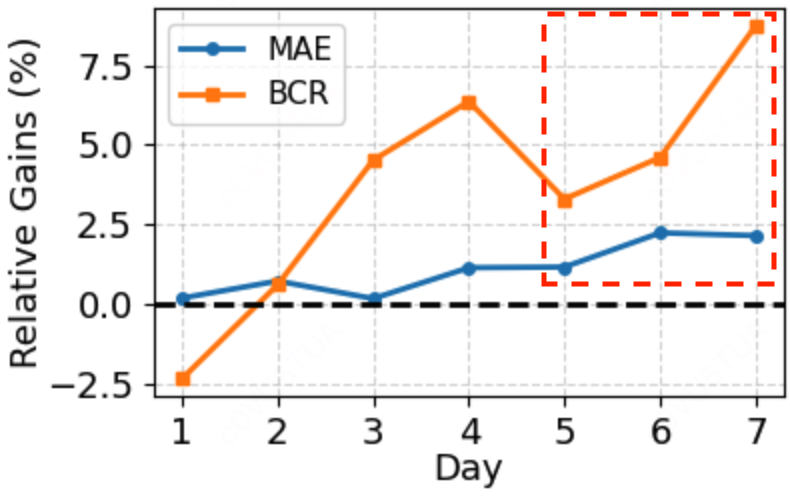}}
\caption{Daily performance gains over control group. Red squares mark greater gains \wrt both metrics in holidays.}
\label{fig:abtest}
\end{figure}

\tabref{tab:abtest} reports the TTE performances of two groups in two metropolises, denoted as City X and City Y, over a period of 1 week, ranging from 2025-04-28 to 2025-05-04 and 2025-05-27 to 2025-06-02, respectively.
We can see that \model consistently outperforms DiDi's current TTE system \wrt MAE and BCR, which justifies the superiority of \model in boosting real-time TTE services.
Moreover, \figref{fig:abtest} shows more significant performance gains during the holidays in both cities. 
This demonstrates not only the importance of integrating link-level modeling to inform underrepresented traffic dynamics, but also the efficacy of \model for long-tail generalization.

\begin{table}[t]
\centering
\caption{Performance gains in online TTE test.}
\begin{tabular}{lcccc}
\toprule
 & \multicolumn{2}{c}{City X} & \multicolumn{2}{c}{City Y} \\
\cmidrule(lr){2-3} \cmidrule(lr){4-5}
& MAE (sec) & BCR (\%) & MAE (sec) & BCR (\%) \\
\midrule
Control & 116.72 & 5.35 & 87.75 & 2.77 \\
Treatment & 113.40 & 5.12 & 86.67 & 2.66 \\
Gain (\%) & \textbf{3.03} & \textbf{4.41} & \textbf{1.24} & \textbf{4.30} \\
\bottomrule
\end{tabular}
\label{tab:abtest}
\end{table}


%% file: relatedwork.tex
\section{Related Work}

\textbf{Travel time estimation.} TTE typically involves OD-based~\cite{murat2018kdd,odoracle2023sigmod} and path-based problem settings~\cite{CompactETAkdd2020,ProbTTEkdd2023}. In this work, we focus on the latter one with the target route specified for estimation.
Existing studies along this line can be primarily categorized into route-centric and link-centric methods.
Route-centric methods model the travel route as a whole and deliver TTE in an end-to-end manner.  
Early statistical methods average travel time of similar historical trajectories for approximate estimation~\cite{TPMFPsigmod2013}. 
Recent deep learning methods leverage sequence models like RNNs~\cite{wdr2018kdd,deeptravel2018ijcai,deeptte2018aaai} and attention mechanisms~\cite{hiereta2022kdd, ieta2023kdd} to capture more complex spatio-temporal correlations among in-route links. 
In contrast, link-centric methods focus on capturing the surrounding traffic dynamics of each in-route link to facilitate more context-aware TTE. 
Some works develop sophisticated Spatio-Temporal Graph Neural Networks~(STGNNs) to estimate link travel times, then sum them for route TTE~\cite{constgat2020kdd,heteta2020kdd,googleeta2021cikm,dueta2022cikm,sthr2022vldb,tteformer2024tits}. 
Others focus on enriching raw features of in-route links with traffic context representations~\cite{compacteta2020kdd,ttpnet2022tkde} or future traffic predictions~\cite{cpmoe2024kdd}.
However, all these methods are still confined to the local traffic contexts, which fail to fully unleash the abundant traffic semantics across the entire urban road networks to enhance the representation of each route.

\textbf{Mixture-of-Experts.}
MoE was first introduced by Jacobs~\etal~\cite{adamoe1991nc} to perform conditional computation for mitigating task interference in neural networks. Shazeer~\etal~\cite{sparsemoe2017iclr} first scaled up MoE to billion-parameter architectures with a sparse gating mechanism. Subsequent advancements in routing mechanism~\cite{softmoe2024iclr,hmora2025iclr}, expert design~\cite{deepseekv3,moeplus2025iclr}, distributed training~\cite{deepspeedmoe2022icml}, and infrastructure~\cite{deepseekv3} further accelerate the large-scale application of MoE in various domains except for NLP, including computer vision~\cite{cvmoe2021nips}, multi-model learning~\cite{multimodalmoe2022nips}, graph learning~\cite{graphmoe2023nips,graphmore2025aaai} and time series analysis~\cite{timemoe2025iclr}.
MoE techniques have also started to gain traction in the traffic domain. ST-MoE~\cite{stmoe2023cikm} and DutyTTE~\cite{dutytte2025aaai} adopt a sparse-gated MoE framework for traffic prediction debiasing and uncertainty quantification, respectively. TESTAM~\cite{testam2024iclr}, CP-MoE~\cite{cpmoe2024kdd} and MH-MoE~\cite{mhmoe} design heterogeneous experts for capturing diverse traffic patterns.
However, these methods only preliminarily testify the conditional computation advantage of MoE in traffic prediction, failing to efficiently and stably handle large-scale road networks with long-tail traffic distributions.

\textbf{Incremental learning.}
Generally, there are three types of IL settings, including Class-Incremental Learning~(CIL), Task-Incremental Learning~(TIL), and Domain-Incremental Learning~(DIL)~\cite{threeil2022nmi}. Since the travel time does not have clear class boundaries, we focus on DIL in this work, which aims to continuously learn from new data distributions while preventing catastrophic forgetting issue.
Existing DIL methods can be categorized into three types: (1) regularization-based ones~\cite{kirkpatrick2017overcoming}, (2) sample replay-based ones~\cite{expreplay2019nips}, and (3) parameter isolation and expansion techniques~\cite{incredeepada2020tpami}. Specifically, in the transportation domain, these three types of DIL approaches have been adapted to address traffic distribution shifts. To name a few, iETA~\cite{ieta2023kdd} adopts self-distillation regularization and periodic trip data replay to mitigate catastrophic forgetting during IL. PECPM~\cite{pecpm2023kdd} and TEAM~\cite{team2024vldb} maintain a representative pattern library to stably adapt to evolving road networks. TFMoE~\cite{tfmoe2024} identifies the most volatile and stable nodes in the road network for selective updates. EAC~\cite{eac2025iclr} introduces lightweight prompts to expand the model's capability of adaptation. In this work, we develop a parameter isolation-based IL strategy tailored for a multi-level TTE framework to support high-frequency adaptation with sustainable computational costs. 

%% file: conclusion.tex
\section{Conclusion}
In this paper, we present \model, a scalable and adaptive framework that synergistically integrates link-level modeling to overcome the limited reception field and long-tail bottleneck of DiDi's current route-centric TTE system. Through the spatio-temporal external attention module and externally stabilized graph MoE, \model efficiently captures global dependencies and heterogeneous traffic patterns across million-scale road networks. A tailored asynchronous incremental learning strategy further enables modular adaptation to frequent traffic shifts with stability and efficiency.
Extensive experiments and the successful deployment in DiDi's ride-hailing platform validate its practical value.
In future work, we plan to extend \model to a foundational framework that fully unlocks the potential of link-level modeling for boosting various prediction and decision-making services beyond TTE.


\section{Acknowledgments}
This work was supported by the National Natural Science Foundation of China (Grant No.~62572417, No.~92370204), National Key~R\&D Program of China (Grant No.~2023YFF0725004), the Guangzhou Basic and Applied Basic Research Program under Grant No.~2024A04J3279, Education Bureau of Guangzhou Municipality, and CCF-DiDi GAIA Collaborative Research Funds.